%% file: main.tex
\documentclass[10pt,twocolumn,letterpaper]{article}
\usepackage{cvpr}              %
\usepackage{times}
\usepackage{epsfig}
\usepackage{graphicx}
\usepackage{amsmath}
\usepackage{amssymb}

\usepackage[table]{xcolor}
\usepackage{subcaption}
\usepackage{multirow}
\usepackage{bbm}
\usepackage{multirow}
\usepackage{algorithm}
\usepackage[noend]{algpseudocode}
\usepackage{tabularx}
\usepackage{makecell}
\usepackage{microtype}
\usepackage{booktabs}
\usepackage{textcomp}
\usepackage{colortbl}
\usepackage{url}
\input{includes}

\input{define}

\usepackage{cuted}

\usepackage[font=small]{caption}
\usepackage[outline]{contour}

\usepackage[pagebackref,breaklinks,colorlinks=true,linkcolor=blue,citecolor=blue,urlcolor=blue]{hyperref}

\usepackage[capitalize]{cleveref}
\crefname{section}{Sec.}{Secs.}
\Crefname{section}{Section}{Sections}
\Crefname{table}{Table}{Tables}
\crefname{table}{Tab.}{Tabs.}

\def\cvprPaperID{2158} %
\def\confName{CVPR}
\def\confYear{2023}

\begin{document}

\title{DynIBaR: Neural Dynamic Image-Based Rendering}

\author{
Zhengqi Li$^1$,
Qianqian Wang$^{1,2}$,
Forrester Cole$^1$,
Richard Tucker$^1$,
Noah Snavely$^1$
\\[0.5em]
$^1$Google Research \ \ \
$^2$Cornell Tech \ \ \ 
\ \ \
}
\maketitle

\begin{strip}
\centering
\includegraphics[width=\textwidth]{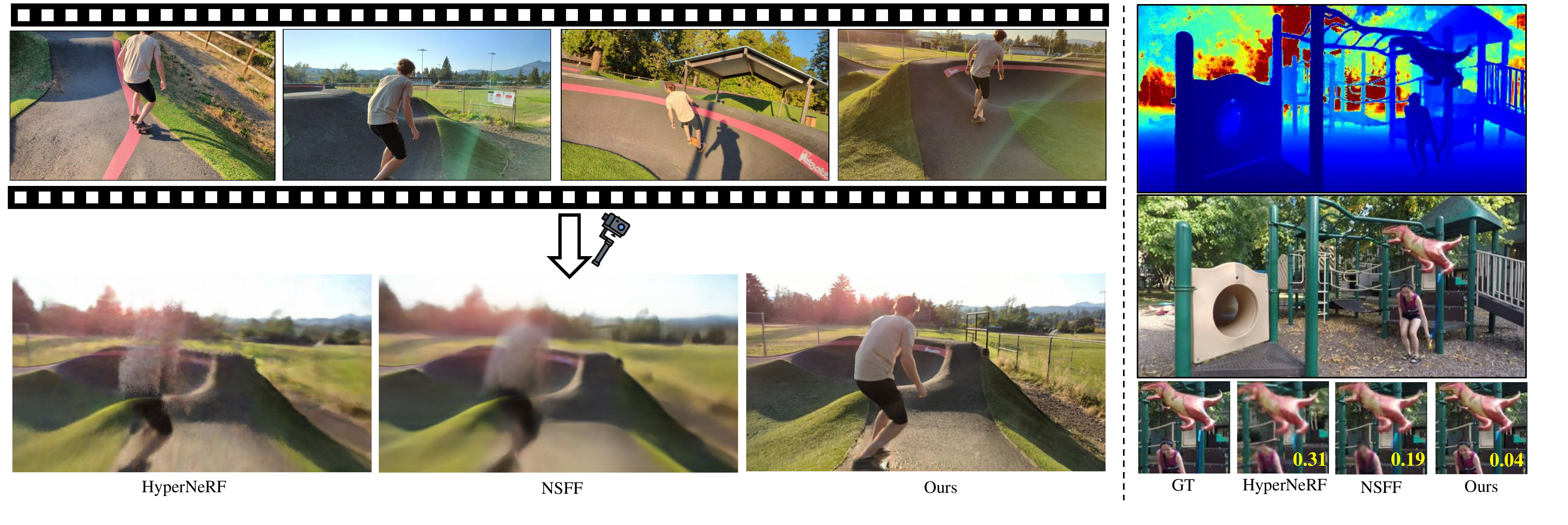} 
\vspace{-2em}
\captionof{figure}{
Recent methods for synthesizing novel views from monocular videos of dynamic scenes--like HyperNeRF~\cite{park2021hypernerf} and NSFF~\cite{li2021neural}--struggle to render high-quality views from long videos featuring complex camera and scene motion. 
We present a new approach that addresses these limitations, illustrated above via an application to $6$DoF video stabilization, where we apply our approach and prior methods on a 30-second, shaky video clip, and compare novel views rendered along a smoothed camera path (\textbf{left}). 
On a dynamic scenes dataset (\textbf{right})~\cite{yoon2020novel}, our approach significantly improves rendering fidelity, as indicated by synthesized images and LPIPS errors computed on pixels corresponding to moving objects (yellow numbers).
Please see the supplementary video for full results.
} \label{fig:teaser}
\end{strip}

\input{abstract}

\input{01-introduction}

\input{02-related}

\input{03-method}

\input{04-evaluation}
\input{05-conclusion}

{\small
\bibliographystyle{ieee_fullname}
\bibliography{refs}
}

\end{document}

%% file: includes.tex
\usepackage{mathalfa}
\usepackage{graphicx}
\usepackage{graphbox}
\usepackage{array}
\usepackage{enumitem}
\usepackage{animate}
\usepackage{xspace} 

\definecolor{rowblue}{RGB}{220,230,240}
\definecolor{myorchid}{RGB}{150,10,30}
\definecolor{myblue}{RGB}{10,30,250}
\definecolor{mygreen}{RGB}{10,190,10}
\definecolor{myred}{RGB}{190,20,20}
\definecolor{mypurple}{RGB}{255,100,255}

\AtBeginDocument{%
 \abovedisplayskip=6pt plus 3pt minus 5pt
 \abovedisplayshortskip=0pt plus 3pt
 \belowdisplayskip=6pt plus 3pt minus 5pt
 \belowdisplayshortskip=5pt plus 3pt minus 4pt
}

\makeatletter
\renewcommand{\paragraph}{%
  \@startsection{paragraph}{4}%
  {\z@}{0.4ex \@plus 1ex \@minus .1ex}{-1em}%
  {\normalfont\normalsize\bfseries}%
}
\makeatother

\makeatletter
\@namedef{ver@everyshi.sty}{} %
\makeatother

\newlength{\itemwidth} %
\usepackage{tikz} %
\usetikzlibrary{calc} %
\usetikzlibrary{tikzmark} %
\usetikzlibrary{spy} %
\usetikzlibrary{shapes.misc} %
\usepackage{pgfplots} %
\usepackage{pgfplotstable} %
\pgfplotsset{compat=newest} %

%% file: define.tex
\newcommand{\Lpho}{\mathcal{L}_{\text{pho}}}

\newcommand{\Ldata}{\mathcal{L}_{\text{data}}}
\newcommand{\Lreg}{\mathcal{L}_{\text{reg}}}

\newcommand{\LMT}{\mathcal{L}_{\text{MT}}}
\newcommand{\Lcp}{\mathcal{L}_{\text{cpt}}}
\newcommand{\Lmask}{\mathcal{L}_{\text{mask}}}
\newcommand{\MaxFrameRange}{r^{\text{max}}}
\newcommand{\NumSrcViews}{N^{\text{vs}}}

\newcommand{\MLP}{G_{\text{MT}}}

\newcommand{\Camera}{\mathbf{P}}

\newcommand{\ray}{\mathbf{r}}

\newcommand{\Pos}{\mathbf{x}} %

\newcommand{\coeff}{\phi} %
\newcommand{\basis}{h} %

\newcommand{\itoj}{{i \rightarrow j}}
\newcommand{\jtok}{{j \rightarrow k}}

\newcommand{\cc}{\mathbf{c}}

\newcommand{\colorStatic}{\mathbf{c}} %
\newcommand{\colorRender}{\hat{\mathbf{C}}} %
\newcommand{\colorGTStatic}{{\mathbf{C}}} %
\newcommand{\feature}{{\mathbf{f}}} %

\newcommand{\colorRenderDy}{\hat{\mathbf{C}}^{\text{dy}}_i} %

\newcommand{\AED}{D} %

\newcommand{\colorj}{\mathbf{c}_j}

\newcommand{\colorRenderUnion}{\hat{\mathbf{C}}^{\text{full}}}

\newcommand{\colorRenderj}{\hat{\mathbf{C}}_{j \rightarrow i}}

\newcommand{\trajectory}{\Gamma}

\newcommand{\rayDisp}{\mathbf{r}_{i \rightarrow j}}

\newcommand{\colorRenderStatic}{\hat{\mathbf{C}}} %

\newcommand{\dynamic}{\text{dy}} %
\newcommand{\static}{\text{st}} %

\newcommand{\alphaED}{\boldsymbol{\alpha}^{\dynamic}_i}
\newcommand{\confidenceED}{\boldsymbol{\beta}^{\dynamic}_i}
\newcommand{\colorED}{\hat{\mathbf{B}}^{\dynamic}_i}

\newcommand{\colorCompose}{\hat{\mathbf{B}}^{\text{full}}_i}
\newcommand{\colorIBR}{\hat{\mathbf{B}}^{\static}}

%% file: abstract.tex
\begin{abstract}
We address the problem of synthesizing novel views from a monocular video depicting a complex dynamic scene. 
State-of-the-art methods based on temporally varying Neural Radiance Fields (aka \emph{dynamic NeRFs}) have shown impressive results on this task.
However, for long videos with complex object motions and uncontrolled camera trajectories, these methods can produce %
blurry or inaccurate renderings, hampering their use in real-world applications.
Instead of encoding the entire dynamic scene within the weights of MLPs, we present a new approach that addresses these limitations by adopting a volumetric image-based rendering framework that synthesizes new viewpoints by aggregating features from nearby views in a scene motion--aware manner.
Our system retains the advantages of prior methods in its ability to model complex scenes and view-dependent effects, but also enables synthesizing photo-realistic novel views from long videos featuring complex scene dynamics with unconstrained camera trajectories.
We demonstrate significant improvements over state-of-the-art methods on dynamic scene datasets, and also 
apply our approach to in-the-wild videos with challenging camera and object motion, where prior methods fail to produce high-quality renderings.
\end{abstract}

%% file: 01-introduction.tex
\section{Introduction}

Computer vision methods can now produce free-viewpoint renderings of static 3D scenes with spectacular quality. What about moving scenes, like those featuring people or pets?
Novel view synthesis from a monocular video of a dynamic scene is a much more challenging \emph{dynamic} scene reconstruction problem. 
Recent work has made progress towards synthesizing novel views in both space and time, thanks to 
new time-varying neural volumetric representations like HyperNeRF~\cite{park2021hypernerf} and Neural Scene Flow Fields (NSFF)~\cite{li2021neural}, which encode spatiotemporally varying scene content volumetrically  
within a coordinate-based multi-layer perceptron (MLP). 

However, these \emph{dynamic NeRF} methods have limitations that prevent their application to casual, in-the-wild videos. 
Local scene flow--based methods like NSFF struggle to scale to longer input videos captured with unconstrained camera motions: the NSFF paper only claims good performance 
for 1-second, forward-facing videos~\cite{li2021neural}. Methods like HyperNeRF that construct a canonical model are mostly constrained to object-centric scenes with controlled camera paths, and can fail on scenes with complex object motion.
 
In this work, we present a new approach that is scalable to dynamic videos captured with 1) long time duration, 2) unbounded scenes, 3) uncontrolled camera trajectories, and 4) fast and complex object motion. Our approach retains the advantages of volumetric scene representations that can model intricate scene geometry with view-dependent effects, while significantly improving rendering fidelity for both static and dynamic scene content compared to recent methods~\cite{li2021neural, park2021hypernerf}, as illustrated in Fig.~\ref{fig:teaser}.

We take inspiration from recent methods for rendering static scenes that synthesize novel images by aggregating local image features from nearby views along epipolar lines~\cite{wang2021ibrnet,suhail2022light, liu2022neuray}. %
However, scenes that are in motion
violate the epipolar constraints assumed by those methods. 
We instead propose to aggregate multi-view image features in \emph{scene motion--adjusted} ray space, which allows us to correctly reason about spatio-temporally varying geometry and appearance.

We also encountered many efficiency and robustness challenges in scaling up aggregation-based methods to dynamic scenes. 
To efficiently model scene motion across multiple views, we model this motion using \emph{motion trajectory fields} that span multiple frames, represented with learned basis functions. 
Furthermore, to achieve temporal coherence in our dynamic scene reconstruction, we introduce a new temporal photometric loss that operates in motion-adjusted ray space.
Finally, to improve 
the quality of novel views, 
we propose to factor the scene into static and dynamic components through a new IBR-based motion segmentation technique within a Bayesian learning framework.

On two dynamic scene benchmarks, we show that our approach 
can render highly detailed scene content and significantly improves upon the state-of-the-art, leading to an average reduction in LPIPS errors by over 50\% both across entire scenes, as well as on regions corresponding to dynamic objects.
We also show that our method can be applied to in-the-wild videos with long duration, complex scene motion, and uncontrolled camera trajectories, where prior state-of-the-art methods fail to produce high quality renderings. We hope that our work advances 
the applicability of dynamic view synthesis methods to real-world videos.

%% file: 02-related.tex
\section{Related Work}

\paragraph{Novel view synthesis.}

Classic image-based rendering (IBR) methods synthesize novel views by integrating pixel information from input images~\cite{shum2000review}, and can be categorized according to their 
dependence on explicit geometry.
Light field or lumigraph rendering methods~\cite{levoy1996light, gortler1996lumigraph,plenopticsampling00,kalantari2016learning} generate new views by filtering and interpolating sampled rays, without use of explicit geometric models. 
To handle sparser input views, many approaches~\cite{debevec1996modeling,buehler2001unstructured,kopf2014first,hedman2016scalable, penner2017soft, hedman2018deep, flynn2016deepstereo,Riegler2020FreeVS,riegler2021stable, kalantari2016learning} leverage pre-computed proxy geometry such as depth maps or meshes to render novel views.

Recently, neural representations have demonstrated 
high-quality novel view synthesis~\cite{zhou2018stereo, srinivasan2019pushing, choi2019extreme, flynn2019deepview, sitzmann2020implicit, sitzmann2019scene, sitzmann2019deepvoxels,lombardi2019neural,mildenhall2020nerf,liu2020neural,niemeyer2020differentiable,wizadwongsa2021nex}. 
In particular, Neural Radiance Fields (NeRF)~\cite{mildenhall2020nerf} achieves an unprecedented level of fidelity by encoding continuous scene radiance fields within multi-layer perceptrons (MLPs). 
Among all methods building on NeRF, IBRNet~\cite{wang2021ibrnet} is the most relevant to our work. 
IBRNet combines classical IBR techniques with volume rendering 
to produce a generalized IBR module that can render high-quality views without per-scene optimization.
Our work extends this kind of volumetric IBR framework designed for static scenes~\cite{wang2021ibrnet,suhail2022light,mvsnerf} to more challenging dynamic scenes. Note that our focus is on synthesizing higher-quality novel views for long videos with complex camera and object motion, rather than on generalization across scenes.

\paragraph{Dynamic scene view synthesis.} 
Our work is related to geometric reconstruction of dynamic scenes from RGBD~\cite{Dou2016Fusion4DRP,newcombe2015dynamicfusion,Innmann2016VolumeDeformRV,zollhofer2014real, bozic2020deepdeform, wang2022neural} or monocular videos~\cite{zhang2021consistent,xuan2020consistent,kopf2021robust,zhang2022structure}. However, depth- or mesh-based representations struggle to model complex geometry and view-dependent effects.

Most prior work on novel view synthesis for dynamic scenes requires multiple synchronized 
input videos~\cite{kanade1997virtualized,bansal20204d,Bemana2020xfields,stich2008view,zitnick2004high,li2022neural,broxton2020immersive, zhang2021editable, wang2022fourier}, limiting their real-world applicability. 
Some methods~\cite{carranza2003free, de2008performance, guo2019relightables,peng2021neural, weng2022humannerf} use domain knowledge such as template models to achieve high-quality 
results, but are restricted to specific categories~\cite{ loper2015smpl,saito2019pifu}.
More recently, many works propose to 
synthesize novel views of dynamic scenes from a single camera. 
Yoon \etal~\cite{yoon2020novel} render novel views 
through explicit warping using depth maps obtained via single-view depth and multi-view stereo. 
However, this method fails to model complex scene geometry and to fill in realistic and consistent content at disocclusions. 
With advances in neural rendering, NeRF-based dynamic view synthesis methods have shown state-of-the-art results~\cite{xian2021space,tretschk2021non,li2021neural,pumarola2021d, du2021neural}. 
Some approaches, such as Nerfies~\cite{park2021nerfies} and HyperNeRF~\cite{park2021hypernerf}, represent scenes using a deformation field mapping each local observation to a canonical scene representation. 
These deformations are conditioned on time~\cite{pumarola2021d} or a per-frame latent code~\cite{tretschk2021non,park2021nerfies,park2021hypernerf}, and are parameterized as translations~\cite{pumarola2021d,tretschk2021non} or rigid body motion fields~\cite{park2021nerfies,park2021hypernerf}.
These methods can handle long videos, but are mostly limited to object-centric scenes with relatively small object motion and controlled camera paths. 
Other methods represent scenes as time-varying NeRFs~\cite{li2021neural,xian2021space,wang2021neural,gao2021dynamic, gao2022monocular}. 
In particular, NSFF uses neural scene flow fields that can capture fast and complex 3D scene motion for in-the-wild videos~\cite{li2021neural}. However, this method only works well for short (1-2 second), forward-facing videos.

%% file: 03-method.tex
\section{Dynamic Image-Based Rendering}

Given a monocular video of a dynamic scene with frames $( I_1, I_2, \ldots, I_N )$ and known camera parameters $(\Camera_1, \Camera_2, \ldots, \Camera_N)$, our goal is to synthesize a novel viewpoint at any desired time within the video. Like many other approaches, we train per-video, first optimizing a model to reconstruct the input frames, then using this model to render novel views.

Rather than encoding 3D color and density directly in the weights of an MLP as in recent dynamic NeRF methods, we integrate classical IBR ideas into a volumetric rendering framework. Compared to explicit surfaces, volumetric representations 
can more readily 
model complex scene geometry with view-dependent effects. 

The following sections introduce our methods for scene-motion-adjusted multi-view feature aggregation (Sec.~\ref{sec:dfa}), and enforcing temporal consistency via \emph{cross-time rendering in motion-adjusted ray space} (Sec.~\ref{sec:pcr}). 
Our full system combines a static model and a dynamic model to produce a color at each pixel. Accurate scene factorization is achieved via segmentation masks derived from a separately trained motion segmentation module within a Bayesian learning framework (Sec.~\ref{sec:mg3d}).

\subsection{Motion-adjusted feature aggregation} \label{sec:dfa}

\begin{figure}[t]
\centering
  \includegraphics[width=1.0\columnwidth]{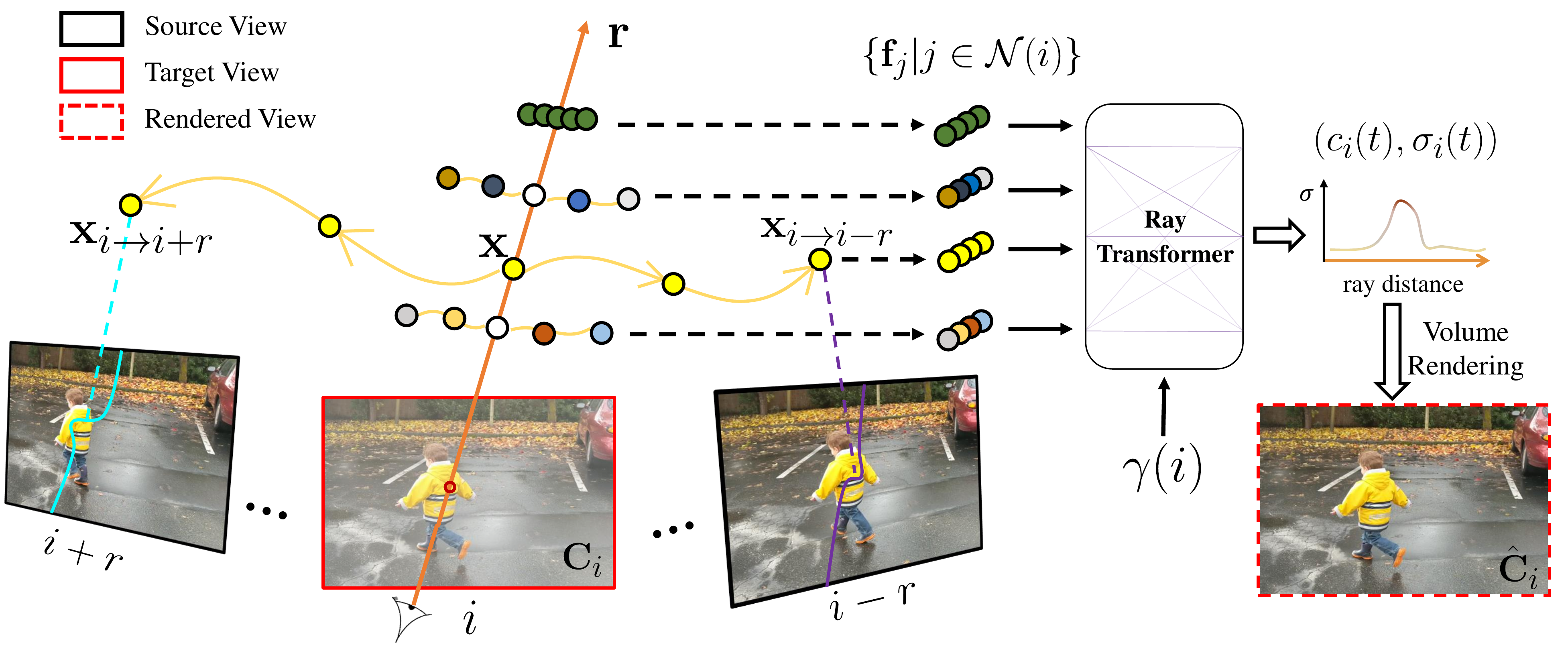} 
  \caption{\textbf{Rendering via motion-adjusted multi-view feature aggregation.} Given a sampled location $\Pos$ at time $i$ along a target ray $\ray$, we estimate its motion trajectory, which determines the 3D correspondence of $\Pos$ at nearby time $j \in \mathcal{N}(i)$, denoted $\Pos_{\itoj}$. Each warped point is then projected into its corresponding source view.
  Image features $\feature_j$ extracted along the projected curves are aggregated and fed to the ray transformer with time embedding $\gamma(i)$, producing per-sample color and density $(\colorStatic_i, \sigma_i)$. The final pixel color $\colorRender_i$ is then synthesized by volume rendering $(\colorStatic_i, \sigma_i)$ along $\ray$.}
\label{fig:overview_1}
\end{figure}

We synthesize new views by aggregating features extracted from temporally nearby source views. To render an image at time $i$, we first identify source views $I_j$ within a temporal radius $r$ frames of $i$, 
$j \in \mathcal{N}(i) = [i - r, i + r]$. For each source view, we extract a 
2D feature map $F_i$ through a shared convolutional encoder network to form an input tuple $\{ I_j, \Camera_j, F_j \}$.

To predict the color and density of each point sampled along a target ray $\ray$, we must aggregate source view features while accounting for scene motion. For a static scene, points along a target ray will 
lie along a corresponding epipolar line in a neighboring source view, hence we can aggregate potential correspondences by simply sampling along neighboring epipolar lines~\cite{wang2021ibrnet, suhail2022light}.
However, moving scene elements violate epipolar constraints, leading to inconsistent feature aggregation if motion is not accounted for. Hence, we perform \emph{motion-adjusted} feature aggregation, as shown in Fig.~\ref{fig:overview_2}.
To determine correspondence 
in dynamic scenes, one straightforward idea is to estimate a scene flow field via an MLP~\cite{li2021neural} to determine a given point's motion-adjusted 3D location at a nearby time. However, this strategy is computational infeasible in a volumetric IBR framework due to recursive unrolling of the MLPs.

\medskip
\noindent \textbf{Motion trajectory fields.}
Instead, we represent scene motion using motion trajectory fields 
described in terms of
learned basis functions. 
For a given 3D point $\Pos$ along target ray $\ray$ at time $i$, we encode its trajectory coefficients with an MLP $\MLP$:
\begin{equation}
\{\coeff^l_i (\Pos) \}_{l=1}^{L} = \MLP \left(\gamma(\Pos), \gamma(i) \right),
\end{equation}
where $\coeff^l_i \in \mathcal{R}^3$ are basis coefficients (with separate coefficients for $x$, $y$, and $z$, using the motion basis described below) and $\gamma$ denotes positional encoding. We choose $L=6$ bases and 16 linearly increasing frequencies for the encoding $\gamma$, based on the assumption that scene motion tends to be low frequency~\cite{zhang2021consistent}. 

We also introduce a global learnable motion basis $\{\basis_i^l\}^{L}_{l=1}, \basis_i^l \in \mathcal{R}$, spanning every time step $i$ of the input video, which is optimized jointly with the MLP. 
The motion trajectory of $\Pos$ is then defined as $\trajectory_{\Pos, i} (j)= \sum_{l=1}^{L} \basis^l_j \coeff^l_i (\Pos)$, and thus, the relative displacement between $\Pos$ and its 3D correspondence $\Pos_{i \rightarrow j}$ at time $j$ is computed as 
\begin{align}
    \Delta_{\Pos, i}(j)= \trajectory_{\Pos, i} (j) - \trajectory_{\Pos, i} (i). \label{eq:relative_displace}
\end{align}
With this motion trajectory representation, finding 3D correspondences for a query point $\Pos$ in neighboring views requires just a single MLP query, allowing efficient multi-view feature aggregation within our volume rendering framework. 
We initialize the basis $\{\basis_i^l\}_{l=1}^L$ with the DCT basis as proposed by Wang~\etal~\cite{wang2021neural}, but fine-tune it along with other components during optimization, since we observe that a fixed DCT basis can fail to model a wide range of real-world motions 
(see third column of Fig.~\ref{fig:ablations}).

Using the estimated motion trajectory of $\Pos$ at time $i$, 
we denote $\Pos$'s corresponding 3D point at time $j$ as $\Pos_\itoj = \Pos + \Delta_{\Pos, i}(j)$.
We project each warped point $\Pos_\itoj$ into its source view $I_j$ 
using camera parameters $\Camera_j$, and extract color and feature vector $\feature_j$ at the projected 2D pixel location. 
The resulting set of source features across neighbor views $j$
is fed to a shared MLP whose output features are aggregated through weighted average pooling~\cite{wang2021ibrnet} to produce a single feature vector at each 3D sample point along ray $\ray$. 
A ray transformer network with time embedding $\gamma(i)$ then processes the sequence of aggregated features along the ray to predict per-sample colors and densities $(\colorStatic_i, \sigma_i)$ (see Fig.~\ref{fig:overview_1}). We then use
standard NeRF volume rendering~\cite{Bi2020NeuralRF} to obtain a final pixel color $\colorRender_i(\ray)$ for the ray from this sequence of colors and densities.

\subsection{Cross-time rendering for temporal consistency} \label{sec:pcr}
\begin{figure}[t]
\centering
  \includegraphics[width=1.0\columnwidth]{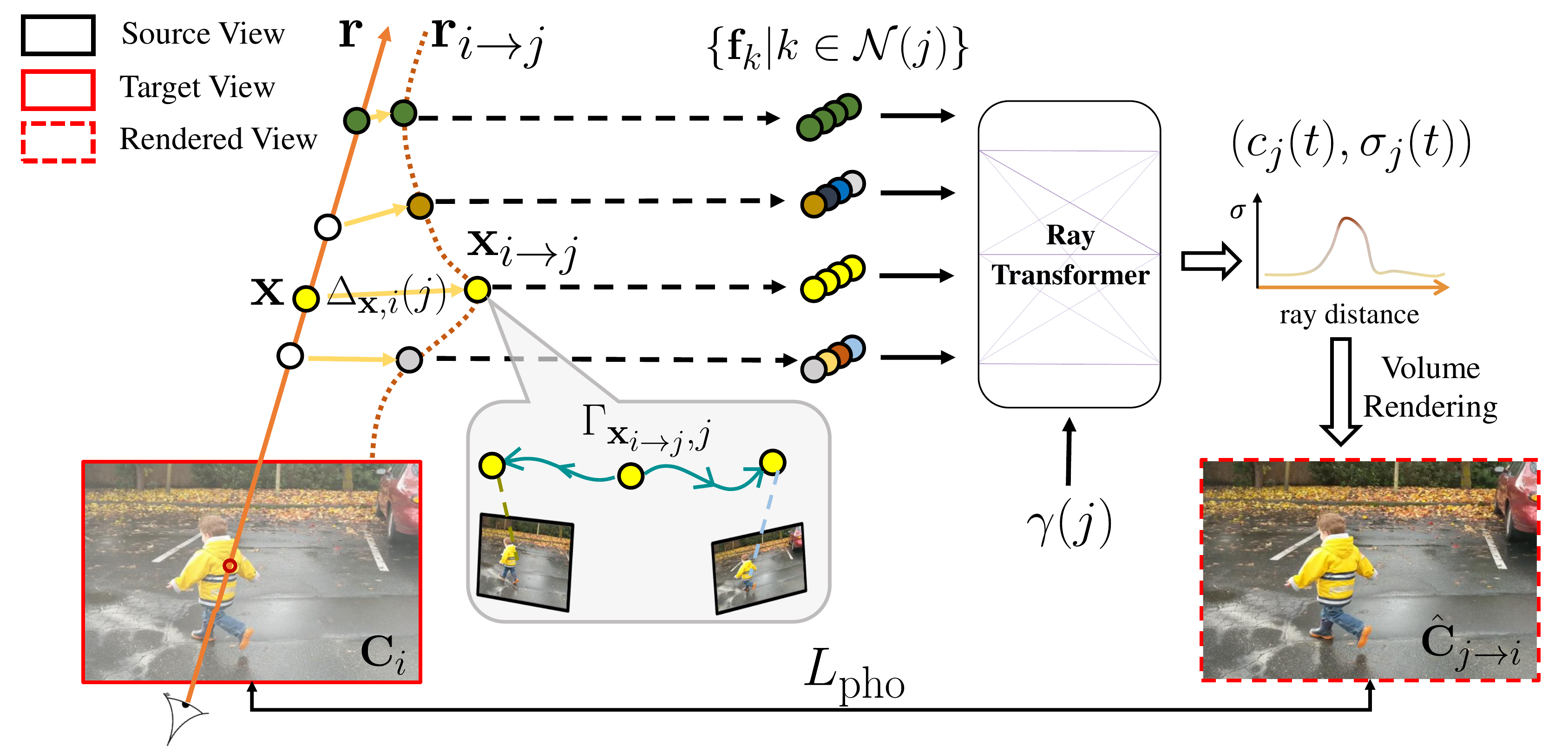} 
  \caption{\textbf{Temporal consistency via cross-time rendering.} To enforce temporal consistency in a dynamic reconstruction, we render each frame $I_i$ using the scene model from a nearby time $j$, which we call \emph{cross-time rendering}.
  A ray $\ray$ from image $i$ is instead rendered using a curved ray $\ray_\itoj$, i.e., $\ray$ warped to time $j$. That is, having computed the motion-adjusted point $\Pos_{i \rightarrow j} = \Pos + \Delta_{\Pos, i}(j)$ at nearby time $j$ from every sampled location along $\ray$, we query $\Pos_{i \rightarrow j}$ and time $j$ via the MLP to predict its motion trajectory $\trajectory_{\Pos_{i \rightarrow j}, j}$, along which we aggregate image features $\feature_k$ extracted from source views within time $k \in \mathcal{N}(j)$. The aggregated features along $\ray_\itoj$ are fed to the ray transformer with time embedding $\gamma(j)$ to produce per-sample color and density $(\colorj, \sigma_j)$ at time $j$. A pixel color $\colorRenderj$ is computed by volume rendering $(\colorj, \sigma_j)$, and then  
  compared to the ground truth color $\colorGTStatic_i$ to form a reconstruction loss $\Lpho$.}
\label{fig:overview_2}
\end{figure}

If we optimize our dynamic scene representation by comparing $\colorRender_i$ with $\colorGTStatic_i$ alone, the representation might overfit to the input images: it might perfectly reconstruct those views, but fail to render correct novel views. 
This can happen because
the representation has the capacity to reconstruct completely separate models for each 
time instance, without utilizing or accurately reconstructing scene motion.
Therefore, to recover a consistent scene with physically plausible motion, we enforce temporal coherence of the scene representation.
One way to define temporal coherence in this context is that the scene at two neighboring times $i$ and $j$ should be consistent when taking scene motion into account\cite{li2021neural, gao2021dynamic, wang2021neural}.

In particular, we enforce temporal photometric consistency in our optimized representation via \emph{cross-time rendering in motion-adjusted ray space}, as shown in Fig.~\ref{fig:overview_2}. 
The idea is to render a view at time $i$ but \emph{via} some nearby time $j$, which we refer to as cross-time rendering. For each nearby time $j \in \mathcal{N}(i)$, rather than directly using points $\Pos$ along ray $\ray$, we consider the points $\Pos_{i \rightarrow j}$ along motion-adjusted ray $\rayDisp$ and treat them as if they lie along a ray at time $j$.%

Specifically, having computed the motion-adjusted points $\Pos_\itoj$, we query the MLP to predict the coefficients of \emph{new} trajectories $\{\coeff^l_j (\Pos_\itoj \}_{l=1}^{L} = \MLP (\Pos_\itoj, \gamma(j))$, and use these to compute corresponding 3D points $(\Pos_\itoj)_\jtok$ for images $k$ in the temporal window $\mathcal{N}(j)$, using Eq.~\ref{eq:relative_displace}.
These new 3D correspondences are then used to render a pixel color exactly as described for a ``straight'' ray $\ray_i$ in Sec.~\ref{sec:dfa}, except now along the curved, motion-adjusted ray $\rayDisp$. 
That is, each point $(\Pos_\itoj)_\jtok$ is projected into its source view $I_k$ and feature maps $F_k$ 
with camera parameters $\Camera_k$ to extract an RGB color and image feature $\feature_k$, and then these features are aggregated and input to the ray transformer with the time embedding $\gamma(j)$. The result is a sequence of colors and densities $(\cc_j, \sigma_j)$ along $\rayDisp$ at time $j$, which can be composited through volume rendering to form a color $\colorRenderj$.

\begin{figure}[t]
\centering
  \includegraphics[width=1.0\columnwidth]{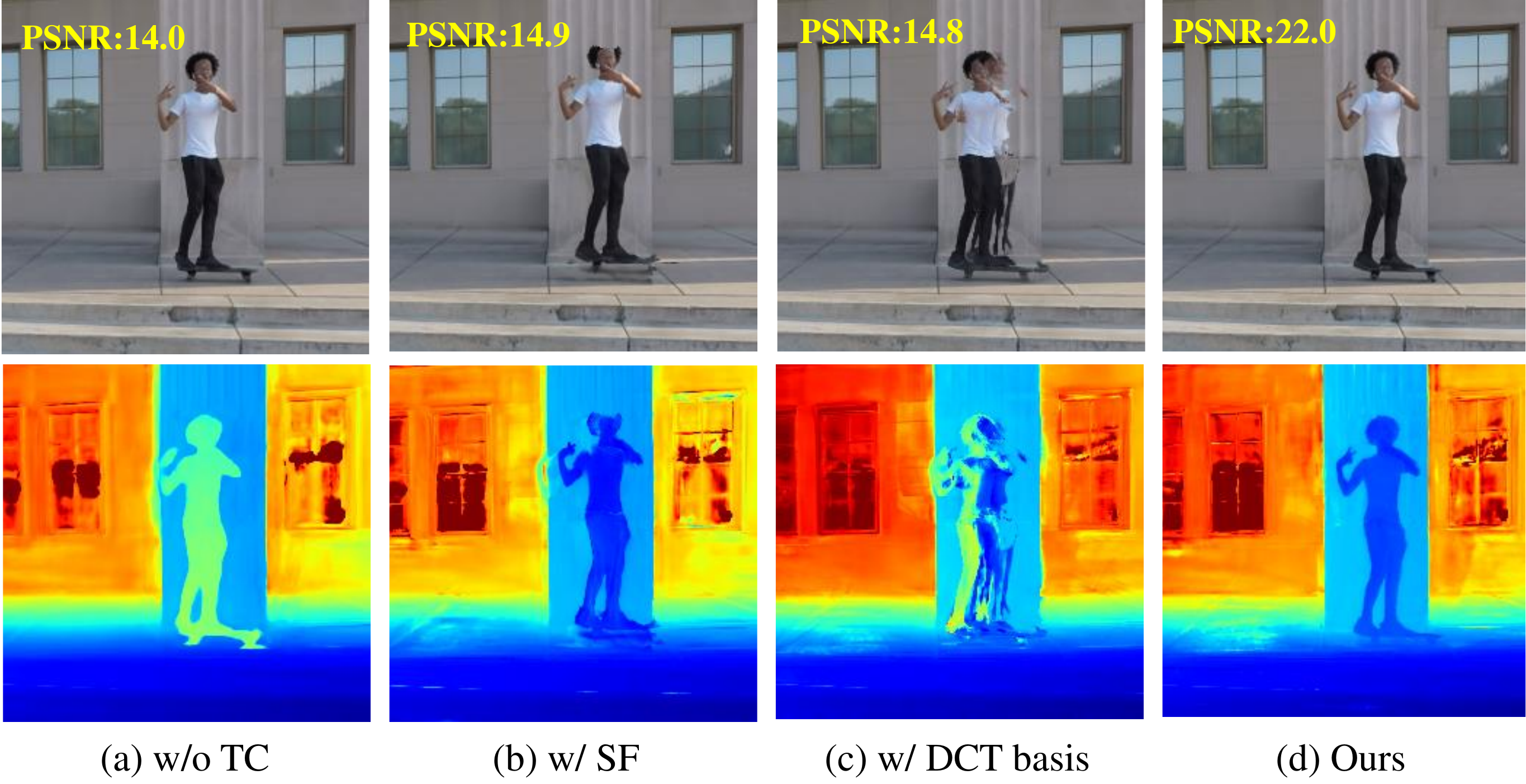} 
  \vspace{-1.5em}
  \caption{\textbf{Qualitative ablations.} From left to right, we show rendered novel views (top) and depths (bottom) from our system (a) without enforcing temporal consistency, (b) aggregating image features with scene flow fields instead of motion trajectories, (c) representing motion trajectory with a fixed DCT basis instead of a learned one, and (d) with full configuration. Simpler configurations significantly degrade rendering quality as indicated by PSNR calculated over the regions of moving objects.}
\label{fig:ablations}
\end{figure}

We can then compare $\colorRenderj (\ray)$ with the target pixel  $\colorGTStatic_i(\ray)$ via a motion-disocclusion-aware 
RGB reconstruction loss:
\begin{align}
    \Lpho = \sum_{\ray} \sum_{j \in \mathcal{N}(i)} \hat{\mathbf{W}}_{j \rightarrow i} (\ray) \rho (\colorGTStatic_i(\ray), \colorRenderj (\ray)). \label{eq:photometric_consistency}
\end{align}
We use a generalized Charbonnier loss~\cite{charbonnier1994two} for the RGB loss $\rho$. $\hat{\mathbf{W}}_{j \rightarrow i} (\ray)$ is a motion disocclusion weight computed by the difference of accumulated alpha weights between time $i$ and $j$ to address the motion disocclusion ambiguity described by NSFF~\cite{li2021neural} (see supplement for more details). Note that when $j = i$, there is no scene motion--induced displacement, meaning $\colorRenderj = \colorRender_i$ and no disocclusion weights are involved ($\hat{\mathbf{W}}_{j \rightarrow i}=1$). 
We show a comparison between our method with and without enforcing temporal consistency in the first column of Fig.~\ref{fig:ablations}.

\subsection{Combining static and dynamic models} \label{sec:mg3d}

As observed in NSFF, synthesizing novel views using a small temporal window is insufficient to recover complete and high-quality content for static scene regions, since the contents may only be observed in spatially distant frames due to uncontrolled camera paths.
Therefore, we follow the ideas of NSFF~\cite{li2021neural}, and model the entire scene using two separate representations. Dynamic content $(\colorStatic_i, \sigma_i)$ is represented with a time-varying model as above (used for cross-time rendering during optimization). 
Static content $(\colorStatic, \sigma)$ is represented with a time-invariant model, which renders in the same way as the time-varying model, but aggregates multi-view features without scene motion adjustment (i.e., along epipolar lines).

The dynamic and static predictions are combined and rendered to a single output color $\colorRenderUnion_i$ (or $\colorRenderUnion_{j \rightarrow i}$ during cross-time rendering) using the method for combining static and transient models of NeRF-W~\cite{MartinBrualla2020NeRFIT}. Each model's color and density estimates can also be rendered separately, giving color $\colorRenderStatic^\static$ for static content and $\colorRenderStatic_i^\dynamic$ for dynamic content.

When combining the two representations, we rewrite the photometric consistency term in Eq.~\ref{eq:photometric_consistency} as a loss comparing $\colorRenderUnion_{j \rightarrow i} (\ray)$ with target pixel $\colorGTStatic_i(\ray)$:
\begin{align}
    \Lpho = \sum_{\ray} \sum_{j \in \mathcal{N}(i)} \hat{\mathbf{W}}_{j \rightarrow i} (\ray) \rho (\colorGTStatic_i(\ray), \colorRenderUnion_{j \rightarrow i} (\ray))
\end{align}

\paragraph{Image-based motion segmentation.}
In our framework, we observed that without any initialization, scene factorization tends to be dominated by either the time-invariant or the time-varying representation, a phenomena also observed in recent methods~\cite{Lu2020LayeredNR, kasten2021layered}.
To facilitate factorization, Gao$~\etal$~\cite{gao2021dynamic} initialize their system 
using masks from semantic segmentation, relying on the assumption that all moving objects are captured by a set of candidate semantic segmentation labels, and segmentation masks are temporally accurate. 
However, these assumptions do not hold in many real-world scenarios, as observed by Zhang~\etal~\cite{casualsam2022}. 
Instead, we propose a new motion segmentation module that produces segmentation masks for supervising our main two-component scene representation. 
Our idea is inspired by the Bayesian learning techniques proposed in recent work~\cite{MartinBrualla2020NeRFIT,casualsam2022}, but
integrated into a volumetric IBR representation for dynamic videos.

In particular, before training our main two-component scene representation,
we jointly train two lightweight models to obtain a motion segmentation mask $M_i$ for each input frame $I_i$. 
We model static scene content with an IBRNet~\cite{wang2021ibrnet} that renders a pixel color $\colorIBR$ using volume rendering along each ray via feature aggregation along epipolar lines from nearby source views without considering scene motion;
we model dynamic scene content with a 2D convolutional encoder-decoder network $\AED$, which predicts a 2D opacity map $\alphaED$, confidence map $\confidenceED$, and RGB image $\colorED$ from an input frame:
\begin{equation}
\colorED, \alphaED, \confidenceED = \AED(I_i).
\end{equation}
The full reconstructed image is then composited pixelwise from the outputs of the two models:
\begin{align}
    \colorCompose (\ray) = \alphaED  (\ray) \colorED  (\ray)
    +  (1 - \alphaED  (\ray)) \colorIBR  (\ray).
\end{align}

To segment moving objects, we assume the observed pixel color is uncertain in a heteroscedastic aleatoric manner, and model the observations in the video with a Cauchy distribution with time dependent confidence $\confidenceED$. By taking the negative log-likelihood of the observations, our segmentation loss is written as a weighted reconstruction loss:
\begin{align}
    \mathcal{L}_{\text{seg}} = \sum_{\ray}\log \left(\confidenceED (\ray) + \frac{|| \colorCompose (\ray) - \colorGTStatic_i (\ray)||^2}{\confidenceED (\ray)}\right). \label{eq:mot_seg}
\end{align}

\begin{figure}[t]
    \centering
    \setlength{\tabcolsep}{0.02cm}
    \setlength{\itemwidth}{2.cm}
    \renewcommand{\arraystretch}{0.5}
    \hspace*{-\tabcolsep}\begin{tabular}{cccc}
            \includegraphics[width=\itemwidth]{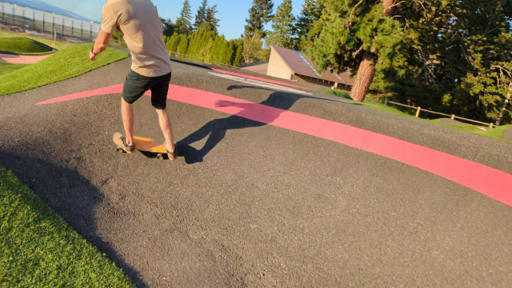} &
            \includegraphics[width=\itemwidth]{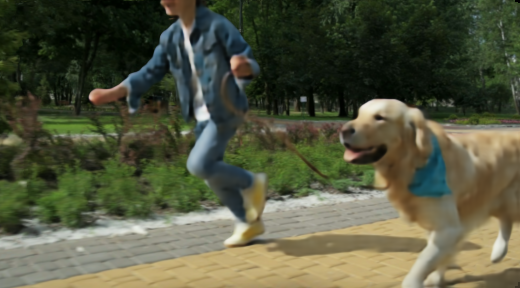} &
            \includegraphics[width=\itemwidth]{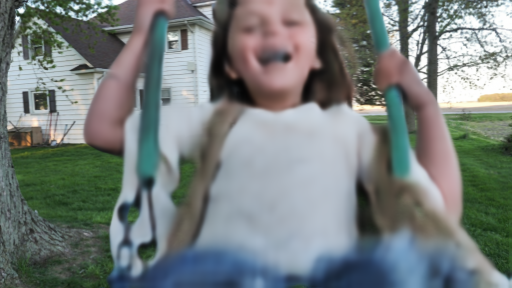} &
            \includegraphics[width=\itemwidth]{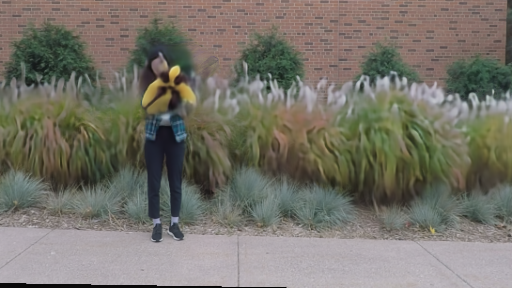} \\
            \includegraphics[width=\itemwidth]{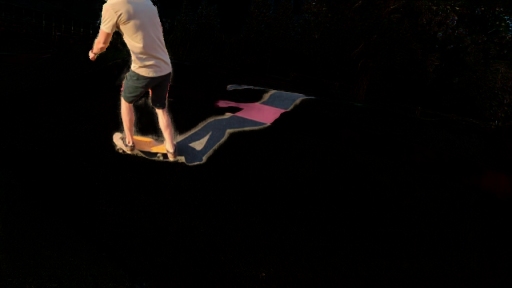} &
            \includegraphics[width=\itemwidth]{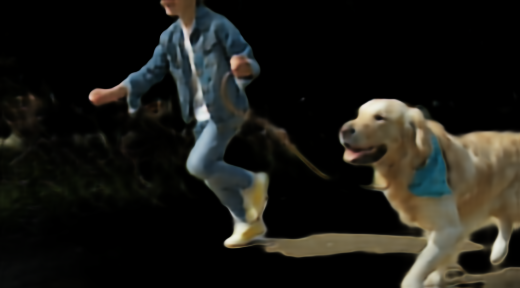} &
            \includegraphics[width=\itemwidth]{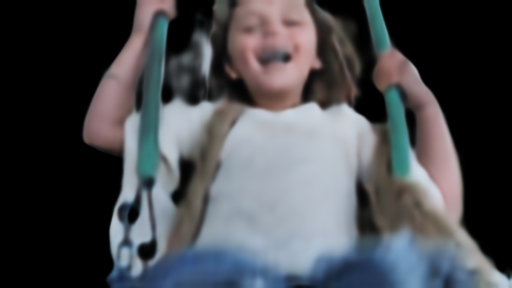} &
            \includegraphics[width=\itemwidth]{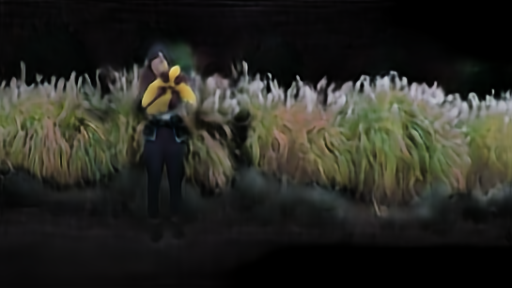} \\
        \\
    \end{tabular}\vspace{-1em}
    \caption{\textbf{Motion segmentation.} We show full rendering $\colorCompose$ (top) and motion segmentation overlaid with rendered dynamic content $\alphaED \odot \colorED$ (bottom). Our approach segments challenging dynamic elements such as the moving shadow, swing, and swaying bushes.}
    \label{fig:render_decomposition}
\end{figure}

By optimizing the two models using Eq.~\ref{eq:mot_seg}, we obtain a motion segmentation mask $M_i$ by thresholding $\alphaED$ at $0.5$. We do not require an alpha regularization loss as in NeRF-W~\cite{MartinBrualla2020NeRFIT} to avoid degeneracies, since we naturally include such an inductive basis by excluding skip connections from network $D$, which leads $D$ to converge more slowly than the static IBR model. We show our estimated motion segmentation masks overlaid on input images in Fig.~\ref{fig:render_decomposition}.

\medskip
\noindent \textbf{Supervision with segmentation masks.}
We initialize our main time-varying and time-invariant models with masks $M_i$ as in Omnimatte~\cite{lu2021omnimatte}, by applying a reconstruction loss to renderings from the time-varying model in dynamic regions, and to renderings from the time-invariant model in static regions:
\begin{align}
    \Lmask = & \sum_{\ray} (1 - M_i) (\ray)  \rho (\colorRenderStatic^\static (\ray), \colorGTStatic_i (\ray)) \nonumber \\
 + & \sum_{\ray}  M_i(\ray) \rho (\colorRenderDy (\ray), \colorGTStatic_i (\ray))
\end{align}
We perform morphological erosion and dilation on $M_i$ to obtain masks of dynamic and static regions respectively in order to turn off the loss near mask boundaries. We supervise the system with $\Lmask$ and decay the weights by a factor of 5 for dynamic regions every 50K optimization steps.

\subsection{Regularization}
As noted in prior work, monocular reconstruction of complex dynamic scenes is highly ill-posed, and using photometric consistency alone is insufficient to avoid bad local minima during optimization~\cite{li2021neural, gao2021dynamic}. 
Therefore, we adopt regularization schemes used in prior work~\cite{barron2022mip, li2021neural, yoon2020novel, wu2022d}, which consist of three main parts $\Lreg = \Ldata + \LMT + \Lcp$. $\Ldata$ is a data-driven term consisting of $\ell_1$ monocular depth and optical flow consistency priors using the estimates from Zhang~\etal~\cite{zhang2021consistent} and RAFT~\cite{teed2020raft}. 
$\LMT$ is a motion trajectory regularization term that encourages estimated trajectory fields to be cycle-consistent and spatial-temporally smooth. 
$\Lcp$ is a compactness prior that encourages the scene decomposition to be binary via an entropy loss, and mitigates floaters through distortion losses~\cite{barron2022mip}. We refer readers to the supplement for more details.

In summary, the final combined loss used to optimize our main representation for space-time view synthesis is:
\begin{align}
    \mathcal{L} & = \Lpho + \Lmask + \Lreg.
\end{align}

%% file: 04-evaluation.tex
\section{Implementation details} 

\paragraph{Data.}
We conduct numerical evaluations on the Nvidia Dynamic Scene Dataset~\cite{yoon2020novel} and UCSD Dynamic Scenes Dataset~\cite{lin2021deep}. Each dataset consists of eight forward-facing dynamic scenes recorded by synchronized multi-view cameras. We follow prior work~\cite{li2021neural} to 
derive a monocular video from each sequence, where each video contains 100$\sim$250 frames. We removed frames that lack large regions of moving objects. We follow the protocol from prior work~\cite{li2021neural} that uses held-out images per time instance for evaluation. We also tested the methods on in-the-wild monocular videos, which feature more challenging camera and object motions~\cite{gao2022monocular}.

\paragraph{View selection.}
For the time-varying, dynamic model 
we use a frame window radius of $r=3$ for all experiments. 
For the time-invariant model representing static scene content, we use separate strategies for the dynamic scenes benchmarks and for in-the-wild videos. 
For the benchmarks, where camera viewpoints are located at discrete camera rig locations, we choose all nearby distinct viewpoints whose timestep is within 12 frames of the target time.
For in-the-wild videos, naively choosing the nearest source views can lead to poor reconstruction due to insufficient camera baseline. 
Thus, to ensure that for any rendered pixel we have sufficient source views for computing its color, we select source views from distant frames. 
If we wish to select $\NumSrcViews$ source views for the time-invariant model, we sub-sample every $\frac{2 \MaxFrameRange}{\NumSrcViews}$ frames from the input video to build a candidate pool, where for a given target time $i$, we only search source views within $[i - \MaxFrameRange, i + \MaxFrameRange ]$ frames. 
We estimate $\MaxFrameRange$ using the method of Li~\etal~\cite{li2019learning} based on SfM point co-visibility and camera relative baseline.
We then construct a final set of source views 
for the model by choosing the top $\NumSrcViews$ frames in the candidate pool that are the closest to the target view in terms of camera baseline. We set $\NumSrcViews=16$.

\paragraph{Global spatial coordinate embedding}
With local image feature aggregation alone, it is hard to determine density accurately on non-surface or occluded surface points due to inconsistent features from different source views, as described in NeuRay~\cite{liu2022neuray}. Therefore, to improve global reasoning for density prediction, we append a global spatial coordinate embedding as an input to the ray transformer, in addition to the time embedding, similar to the ideas from~\cite{suhail2022light}. Please see supplement for more details.

\paragraph{Handling degeneracy through virtual views.}
Prior work~\cite{li2021neural} observed that optimization can converge to bad local minimal if camera and object motions are mostly colinear, or scene motions are too fast to track. Inspired by ~\cite{li2022infinitenature}, we synthesize images at eight randomly sampled nearby viewpoints for every input time via depths estimated by~\cite{zhang2021consistent}. During rendering, we randomly sample virtual view as additional source image. We only apply this technique to in-the-wild videos since it help avoid degenerate solutions while improving rendering quality, whereas we don't observe improvement on the benchmarks due to disentangled camera-object motions as described by~\cite{gao2022monocular}.

\paragraph{Time interpolation.}
Our approach also allows for time interpolation by performing scene motion--based splatting, as introduced by NSFF~\cite{li2021neural}.
To render at a specified target fractional time, we predict the volumetric density and color at two nearby input times by aggregating local image features from their corresponding set of source views. The predicted color and density are then splatted and linearly blended via the scene flow derived from our motion trajectories, and weighted according to the target fractional time index.

\begin{table}[tb]
\resizebox{\columnwidth}{!}{%
\begin{tabular}{lcccccc} 
\toprule
\multirow{2}{*}{Methods}
& \multicolumn{3}{c}{Full} & \multicolumn{3}{c}{Dynamic Only} \\
\cmidrule(lr){2-4} \cmidrule(lr){5-7} 
& SSIM$\uparrow$ & PSNR$\uparrow$ & LPIPS$\downarrow$ & SSIM$\uparrow$ & PSNR$\uparrow$ & LPIPS$\downarrow$ \\ 
\midrule
Nerfies~\cite{park2021nerfies} & 0.609 & 20.64 & 0.204 &	0.455 & 17.35 &	0.258 \\
HyperNeRF~\cite{park2021hypernerf} & 0.654 & 20.90 & 0.182 & 0.446 & 17.56 & 0.242  \\
DVS~\cite{gao2021dynamic} & 0.921 & 27.44 & 0.070 & 0.778 & 22.63 & \underline{0.144} \\
NSFF~\cite{li2021neural} & \underline{0.927} & \underline{28.90} & \underline{0.062} & \underline{0.783} & \underline{23.08} & 0.159 \\
\midrule 
Ours & \textbf{0.957} & \textbf{30.86} & \textbf{0.027} & \textbf{0.824} & \textbf{24.24} & \textbf{0.062} \\
\bottomrule
\end{tabular}
} \vspace{-0.5em}
\caption{\textbf{Quantitative evaluation on the Nvidia dataset~\cite{yoon2020novel}.}} \label{exp:nvidia_quan} %
\end{table}

\begin{figure}[tb]
\centering
  \includegraphics[width=1.0\columnwidth]{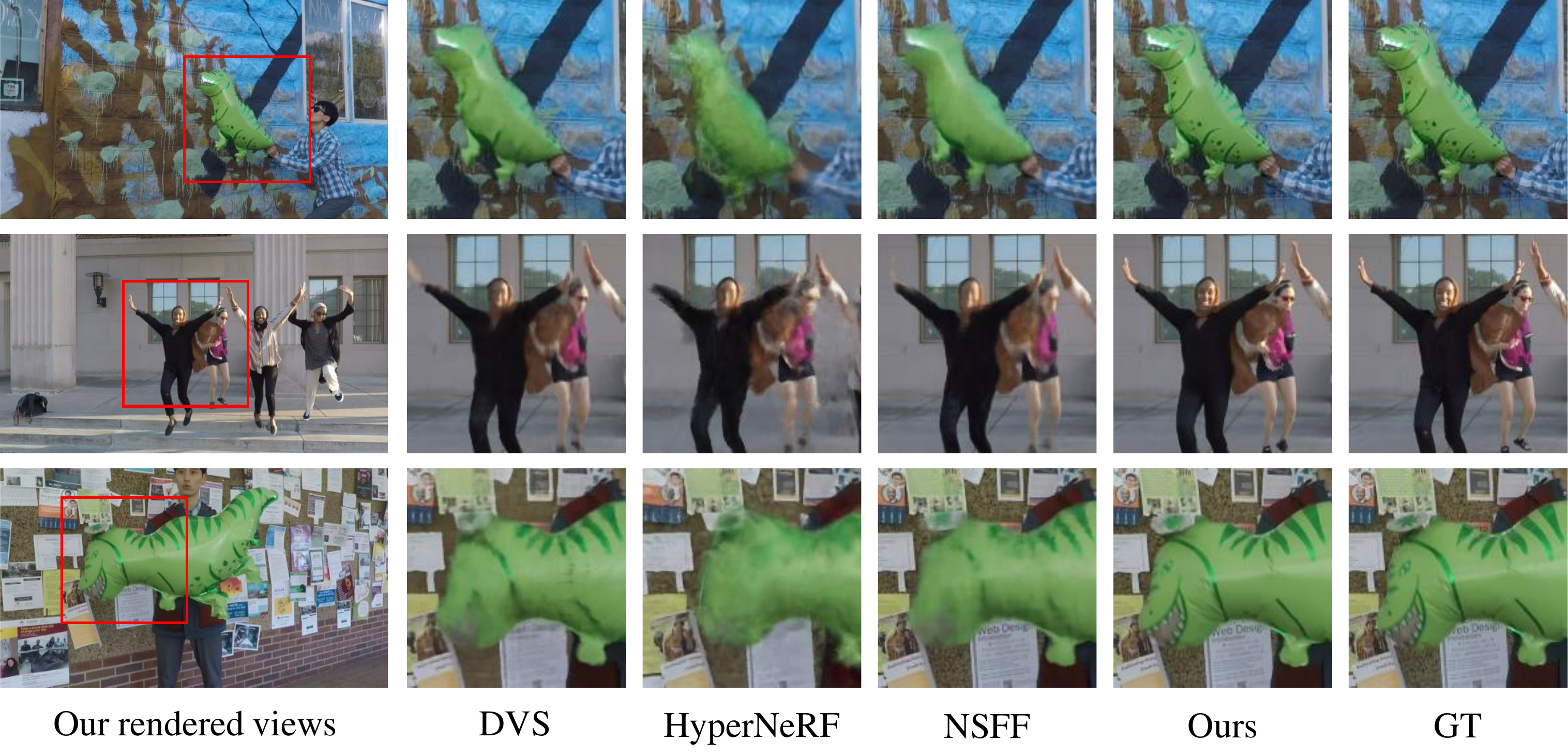} 
    \vspace{-1.5em}
  \caption{\textbf{Qualitative comparisons on the Nvidia dataset~\cite{yoon2020novel}.} }
\label{fig:nvidia_comp}
\end{figure}

\paragraph{Setup.} 
We estimate camera poses using COLMAP~\cite{schonberger2016structure}.
For each ray, we use a coarse-to-fine sampling strategy with 128 per-ray samples as in Wang~\etal~\cite{wang2021ibrnet}. A separate model is trained from scratch for each scene using the Adam optimizer~\cite{Kingma2014AdamAM}. 
The network architecture for two main representations is a variant of that of IBRNet~\cite{wang2021ibrnet}. We reconstruct the entire scene in Euclidean space without special scene parameterization. 
Optimizing a full system on a 10 second video takes around two days using 8 Nvidia A100s, and rendering takes roughly 20 seconds for a $768 \times 432$ frame. We refer readers to the supplemental material for network architectures, hyper-parameters setting, and additional details.

\section{Evaluation}

\subsection{Baselines and error metrics}
We compare our approach to state-of-the-art monocular view synthesis methods.
Specifically, we compare to three recent canonical space--based methods, Nerfies~\cite{park2021nerfies},  and HyperNeRF~\cite{park2021hypernerf}, and to two scene flow--based methods, NSFF~\cite{li2021neural} and Dynamic View Synthesis (DVS) from Gao$~\etal$~\cite{gao2021dynamic}. For fair comparisons, we use the same depth, optical flow and motion segmentation masks used for our approach as inputs to other methods.

Following prior work~\cite{li2021neural}, we report the rendering quality of each method with three standard error metrics: peak signal-to-noise ratio (PSNR), structural similarity (SSIM), and perceptual similarity via LPIPS~\cite{zhang2018unreasonable}, and calculate errors both over the entire scene (Full) and restricted to moving regions (Dynamic Only).

\subsection{Quantitative evaluation}
\label{sec:quant_eval}

Quantitative results on the two benchmark datasets are shown in Table~\ref{exp:nvidia_quan} and Table~\ref{exp:ucsd_numerical}. Our approach significantly improves over prior state-of-the-art methods in terms of all error metrics. Notably, our approach improves PSNR over entire scene upon the second best methods by 2dB and 4dB on each of the two datasets. Our approach also reduces LPIPS error, a major indicator of perceptual quality compared with real images~\cite{zhang2018unreasonable}, by over $50\%$. These results suggest that our framework is much more effective at recovering highly detailed scene contents.

\begin{table}[t]
\resizebox{\columnwidth}{!}{%
\begin{tabular}{lcccccc} 
\toprule
\multirow{2}{*}{Methods}
& \multicolumn{3}{c}{Full} & \multicolumn{3}{c}{Dynamic Only} \\
\cmidrule(lr){2-4} \cmidrule(lr){5-7} 
& SSIM$\uparrow$ & PSNR$\uparrow$ & LPIPS$\downarrow$ & SSIM$\uparrow$ & PSNR$\uparrow$ & LPIPS$\downarrow$ \\ 
\midrule
Nerfies~\cite{park2021nerfies} & 0.823 & 24.32 &	0.096 & 0.595 & 18.45 & 0.234 \\
HyperNeRF~\cite{park2021hypernerf} & 0.859 & 25.10 & 0.095 & 0.618 & 19.26 & 0.212 \\
DVS~\cite{gao2021dynamic} & 0.943 & 30.64 & 0.075 & \underline{0.866} & \underline{26.57} & \underline{0.096} \\
NSFF~\cite{li2021neural} & \underline{0.952} & \underline{31.75} & \underline{0.034} & 0.851 & 25.83 & 0.115 \\
\midrule
Ours  & \textbf{0.983} & \textbf{36.47} & \textbf{0.014} & \textbf{0.909} & \textbf{28.01} & \textbf{0.042} \\\bottomrule
\end{tabular}
} \vspace{-0.5em}
\caption{\textbf{Quantitative evaluation on the UCSD dataset~\cite{lin2021deep}.} \label{exp:ucsd_numerical}}
\end{table}

\begin{figure}[t]
\centering
  \includegraphics[width=1.0\columnwidth]{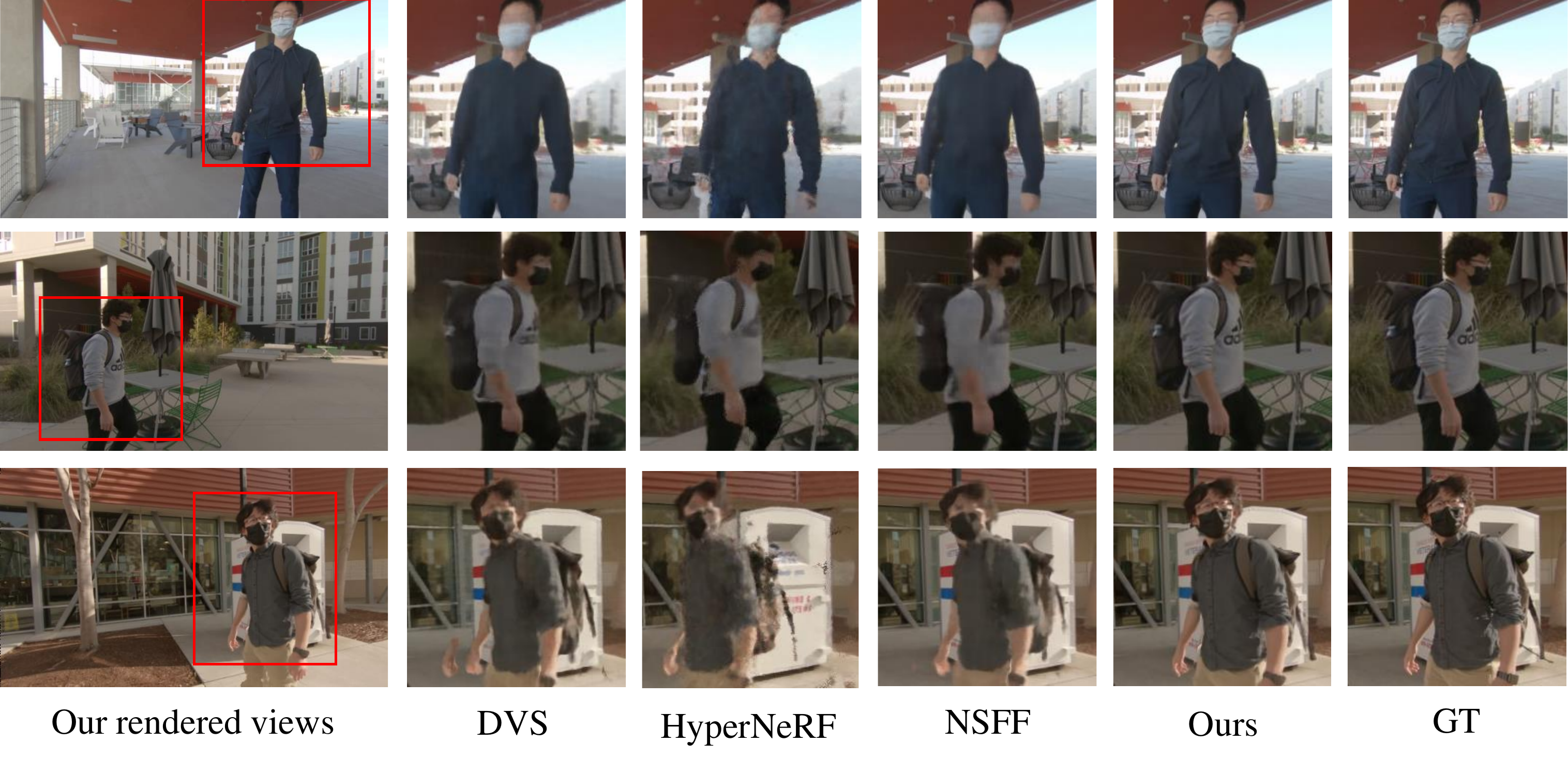} \vspace{-1.5em}
  \caption{\textbf{Qualitative comparisons on the UCSD dataset~\cite{lin2021deep}.} }
\label{fig:ucsd_comp}
\end{figure}

\begin{figure*}[tb]
\centering
  \includegraphics[width=2.1\columnwidth]{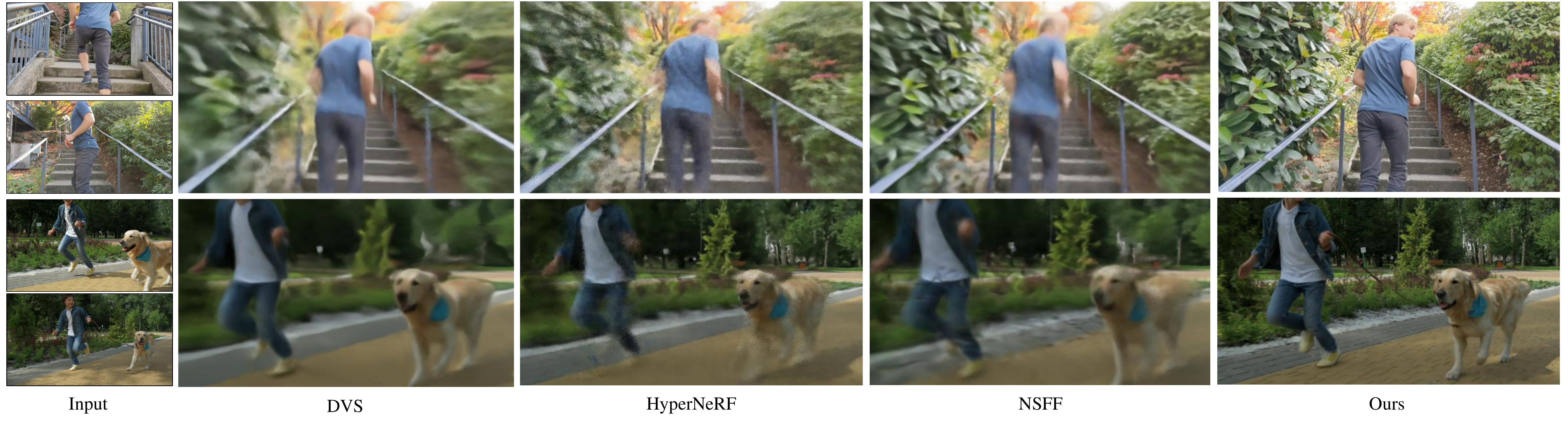} \vspace{-1.5em}
  \caption{\textbf{Qualitative comparisons on in-the-wild videos.} We show results on 10-second videos of complex dynamic scenes. The leftmost column shows the start and end frames of each video; on the right we show novel views at intermediate times rendered from our approach and prior state-of-the-art methods~\cite{gao2021dynamic, park2021hypernerf, li2021neural}. }
\label{fig:viw}
\end{figure*}

\paragraph{Ablation study.}  \label{sec:ablation}
We conduct an ablation study on the Nvidia Dynamic Scene Dataset to validate the effectiveness of our various proposed system components. 
We show comparisons between our full system and variants in Tab.~\ref{exp:ablation_nvidia}: 
A) baseline IBRNet~\cite{wang2021ibrnet} with extra time embedding;
B) without enforcing temporal consistency via cross-time rendering; C) using scene flow fields to aggregate image features within one time step;
D) predicting multiple 3D scene flow vectors pointing to $2r$ nearby times at each sample;
E) without using a time-invariant static scene model;
F) without masked reconstruction loss via estimated motion segmentation masks;
and G) without regularization loss. 
For this ablation study, we train each model 
with 64 samples per ray. Without our motion trajectory representation and temporal consistency, view synthesis quality degrades significantly as shown in the first three rows of Tab.~\ref{exp:ablation_nvidia}. Integrating a global spatial coordinate embedding further improves rendering quality. Combining static and dynamic models improves quality for static elements, as seen in metrics over full scenes. Finally, removing supervision from motion segmentation or regularization reduces overall rendering quality, demonstrating the value of proposed losses for avoiding bad local minimal during optimization.

\subsection{Qualitative evaluation}

\paragraph{Dynamic scenes dataset.} 
We provide qualitative comparisons between our approach and three prior state-of-the-art methods~\cite{gao2021dynamic, li2021neural, park2021hypernerf} on the test views from two datasets in Fig.~\ref{fig:nvidia_comp} and Fig.\ref{fig:ucsd_comp}.
Prior dynamic-NeRF methods have difficulty rendering details of moving objects, as seen in the excessively blurred dynamic content including the texture of balloons, human faces, and clothing. In contrast, our approach synthesizes photo-realistic novel views of both static and dynamic scene content and which are closest to the ground truth images.

\begin{table}[tb]
\resizebox{\columnwidth}{!}{%
\begin{tabular}{lccccccccc} 
\toprule
Methods
& \multicolumn{3}{c}{Full} & \multicolumn{3}{c}{Dynamic Only} \\
\cmidrule(lr){2-4} \cmidrule(lr){5-7} 
& SSIM$\uparrow$ & PSNR$\uparrow$ & LPIPS$\downarrow$ & SSIM$\uparrow$ & PSNR$\uparrow$ & LPIPS$\downarrow$ \\ 
\midrule
A)~\cite{wang2021ibrnet}+time & 0.905 &             25.33 &             0.081 & 0.683 & 20.09	& 0.122  \\
B)~w/o TC &                     0.911 &             27.57 &             0.074 & 0.751 & 22.16	& 0.104  \\ %
C)~w/ SF &  0.935 &             29.42 &             0.035 & 0.797 & 22.41	& 0.095  \\ %
D)~w/ M-SF & 0.947 & 29.59 & 0.033 & 0.814 & 22.97 & 0.084  \\ %
E)~w/o static rep.  &           0.919 &             28.19 &             0.047 & 	\textbf{0.840}	& \text{24.01} &	\text{0.071}  \\
F)~w/o $\Lmask$   &             0.930 &             29.95 &             0.036 & {0.835}	& \textbf{24.30} & \textbf{0.063} \\
G)~w/o $\Lreg$  &               0.921 &             29.46 &	            0.042 & 0.795 & 22.19	& 0.080   \\
\midrule
~Full & \textbf{0.957} & \textbf{30.77} & \textbf{0.028} & \underline{0.837} & \underline{24.27} & \underline{0.066}   \\
\bottomrule
\end{tabular}
} \vspace{-0.5em}
\caption{\textbf{Ablation study on the Nvidia Dataset.} See Sec.~\ref{sec:ablation} for detailed descriptions of each configuration.}
\label{exp:ablation_nvidia}
\end{table}

\begin{figure}[t]
    \centering
    \setlength{\tabcolsep}{0.01cm}
    \setlength{\itemwidth}{2.75cm}
    \renewcommand{\arraystretch}{0.2}
    \hspace*{-\tabcolsep}\begin{tabular}{ccc}
            \includegraphics[width=\itemwidth]{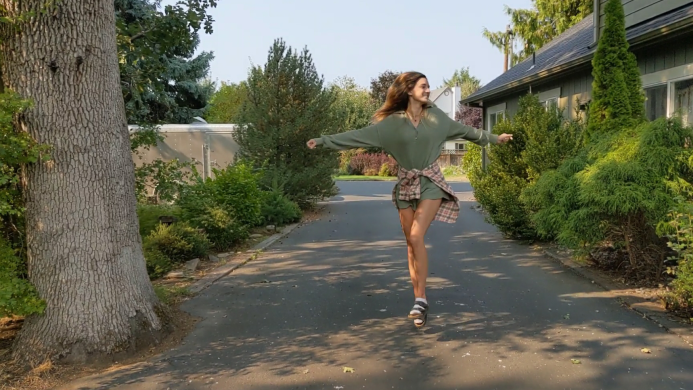} &
            \includegraphics[width=\itemwidth]{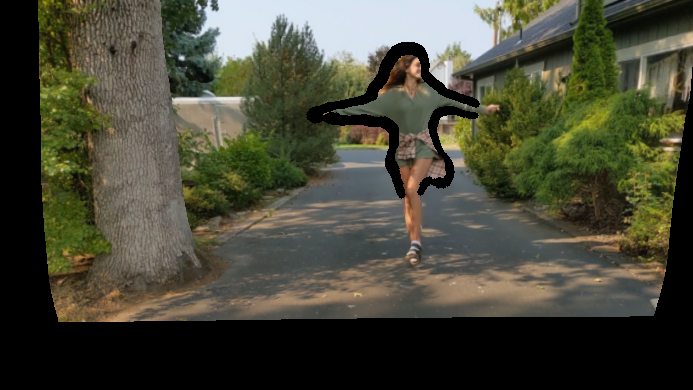} &
            \includegraphics[width=\itemwidth]{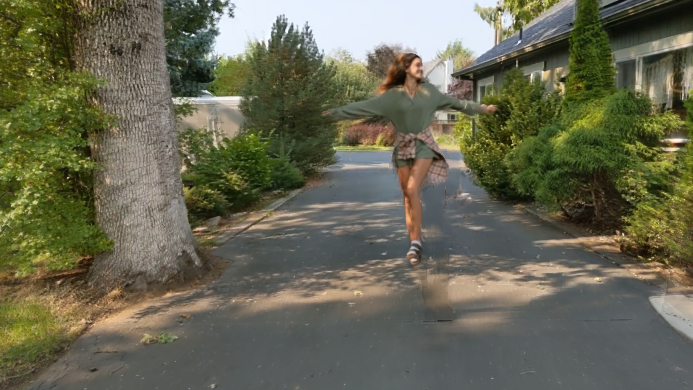} \\
            \includegraphics[width=\itemwidth]{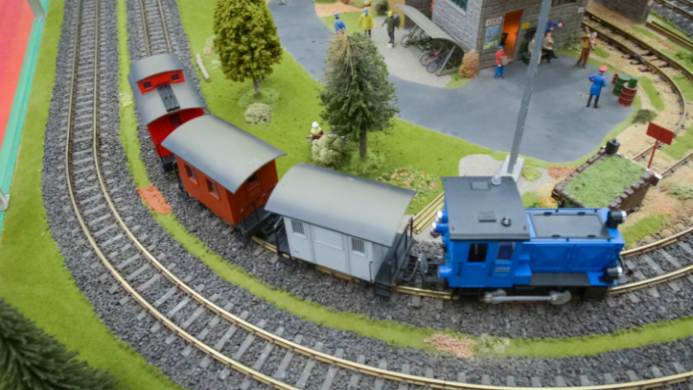}&
            \includegraphics[width=\itemwidth]{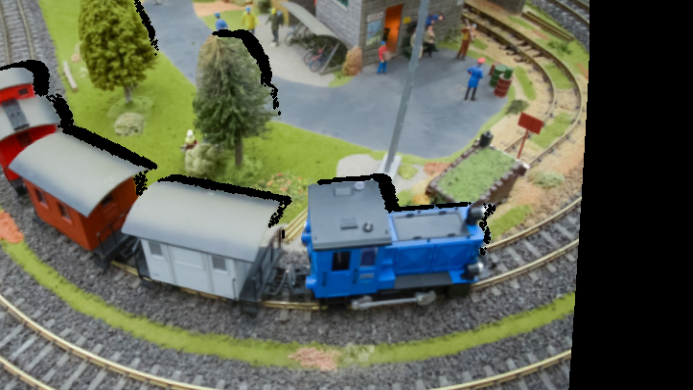} &
            \includegraphics[width=\itemwidth]{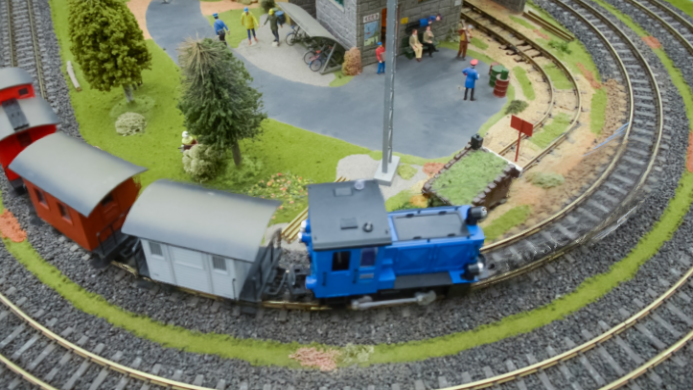} \\
    \end{tabular}\vspace{-0.5em}
    \caption{From left to right, We show inputs and corresponding novel views rendered from explicit depth warping with Zhang~\etal~\cite{zhang2021consistent},  and from our approach.}
    \label{fig:depth_comp} %
\end{figure}

\paragraph{In-the-wild videos.} 
We show qualitative comparisons on in-the-wild footage of complex dynamic scenes. We show comparisons with Dynamic-NeRF based methods in Fig.~\ref{fig:viw}, and Fig.~\ref{fig:depth_comp} shows comparisons with point cloud rendering using depths~\cite{zhang2021consistent} . Our approach synthesizes photo-realistic novel views, whereas prior dynamic-Nerf methods fail to recover high-quality details of both static and moving scene contents, such as the shirt wrinkles and the dog's fur in Fig.~\ref{fig:viw}. On the other hand, explicit depth warping produces holes at regions near disocculusions and out of field of view.  We refer readers to the supplementary video for full comparisons.

%% file: 05-conclusion.tex
\section{Discussion and conclusion}

\begin{figure}[t]
    \centering
    \setlength{\tabcolsep}{0.025cm}
    \setlength{\itemwidth}{2.75cm}
    \renewcommand{\arraystretch}{0.5}
    \hspace*{-\tabcolsep}\begin{tabular}{ccc}
            \includegraphics[width=\itemwidth]{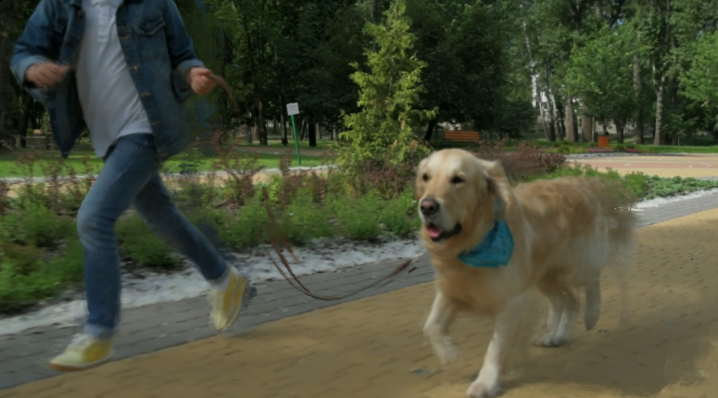} & 
            \includegraphics[width=\itemwidth]{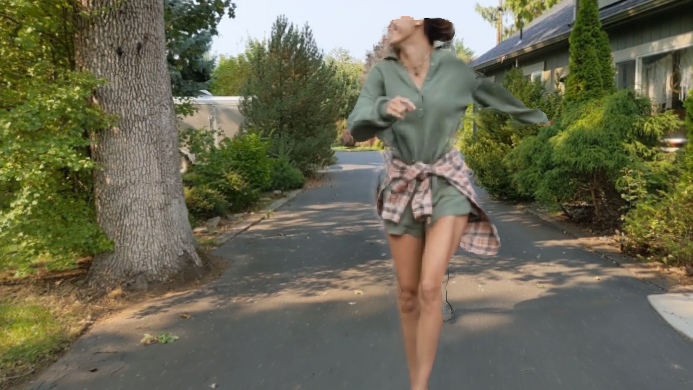} &
            \includegraphics[width=\itemwidth]{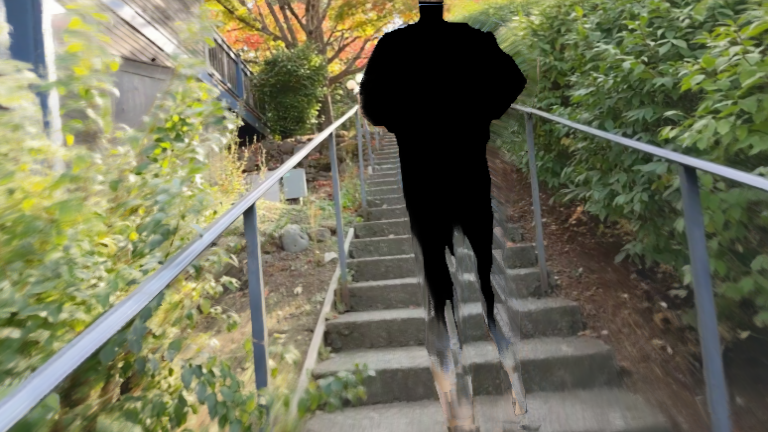} \\
    \end{tabular} %
    \caption{\textbf{Limitations.} Our method might fail to model moving thin objects such as moving leash (left). Our method can fail to render dynamic contents only visible in distant frames (middle). The rendered static content can be unrealistic or blank if insufficient source views feature are aggregated for a given pixel (right).}
    \label{fig:limitations}
\end{figure}

\paragraph{Limitations.} 
 Our method is limited to relatively small viewpoint changes compared with methods designed for static or quasi-static scene; Our method is not able to handle small fast moving object due to incorrect initial depth and optical flow estimates (left, Fig.~\ref{fig:limitations}). In addition, compared to prior dynamic NeRF methods, the synthesized views are not strictly multi-view consistent, and rendering quality of static content depends on which source views are selected (middle, Fig.~\ref{fig:limitations}). Our approach is also sensitive to degenerate motion patterns from in-the-wild videos, in which object and camera motion is mostly colinear, but we show heuristics to handle such cases in the supplemental. Moreover, our method is able to synthesize dynamic contents only appearing at distant time (right, Fig.~\ref{fig:limitations})

\paragraph{Conclusion.}
We presented a new approach for space-time view synthesis from a monocular video depicting a complex dynamic scene. By representing a dynamic scene within a volumetric IBR framework, our approach overcomes limitations of recent methods that cannot model long videos with complex camera and object motion.
We have shown that our method can synthesize photo-realistic novel views from in-the-wild dynamic videos, and can achieve significant improvements over prior state-of-the-art methods on the dynamic scene benchmarks. %

\paragraph{Acknowledgements.}
Thanks to Andrew Liu, Richard Bowen and Lucy Chai for the fruitful discussions, and thanks to Rick Szeliski and Ricardo Martin-Brualla for helpful proofreading.